\documentclass[letterpaper, 10 pt, conference]{ieeeconf}  

\IEEEoverridecommandlockouts                              

\makeatletter
\def\endthebibliography{%
	\def\@noitemerr{\@latex@warning{Empty `thebibliography' environment}}%
	\endlist
}
\makeatother

\usepackage{graphics} 
\usepackage{epsfig} 
\usepackage{mathptmx} 
\usepackage{times} 
\usepackage{amsmath} 
\usepackage{amssymb}  
\usepackage{siunitx}
\usepackage{verbatim}
\usepackage{mathrsfs}
\usepackage{booktabs}
\usepackage[ruled,vlined]{algorithm2e} 

\title{\LARGE \bf  Navigation of a Self-Driving Vehicle Using One Fiducial Marker}

\author{Yibo Liu, Hunter Schofield, Jinjun Shan
\thanks{The authors are with Department of Earth and Space Science and Engineering, York University, Toronto, Ontario M3J1P3, Canada 
        {\tt\footnotesize \{yorklyb,hunterls,jjshan\}@yorku.ca}}%
}

\begin{document}

\maketitle
\thispagestyle{empty}
\pagestyle{empty}

\begin{abstract}
Navigation using only one marker, which contains four artificial features, is a challenging task since camera pose estimation using only four coplanar points suffers from the rotational ambiguity problem in a real-world application. This paper presents a framework of vision-based navigation for a self-driving vehicle equipped with multiple cameras and a wheel odometer. A multiple camera setup is presented for the camera cluster which has $360^{\circ}$ vision such that our framework solely requires one planar marker. A Kalman-Filter-based fusion method is introduced for the multiple-camera and wheel odometry. Furthermore,  an algorithm is proposed to resolve the rotational ambiguity problem using the prediction of the Kalman Filter as additional information.  Finally, the lateral and longitudinal controllers are provided. Experiments are conducted to illustrate the effectiveness of the theory.

\end{abstract}

\section{INTRODUCTION}
Fiducial marker systems, such as AprilTag \cite{olson,wang}, are designed to enhance augmented reality, evaluate performance of robot systems, and improve human-robot interactions, etc. Markers provide the environment with controllable and stable artificial features. This simplifies feature extraction \cite{ch2020}, which is why markers are widely used for tasks that require high detection speed, such as visual servoing \cite{borowczyk}. As markers are being designed to be easily detected from a wider range of locations, planar marker-based SLAM \cite{munoz,munoz2019} is becoming a hot trend. It is fundamental to estimate the 6-DOF pose of the camera with respect to the world or marker coordinate system in research that employs marker-based localization. In practice, the pose is computed using the four corners of the marker. Since the corners are co-planar, this is known as a specific case of  the Perspective-N-Point problem called planar pose estimation \cite{ch2020}.\par
\begin{figure}[thpb]
	\centering
	\includegraphics[width=2.0in]{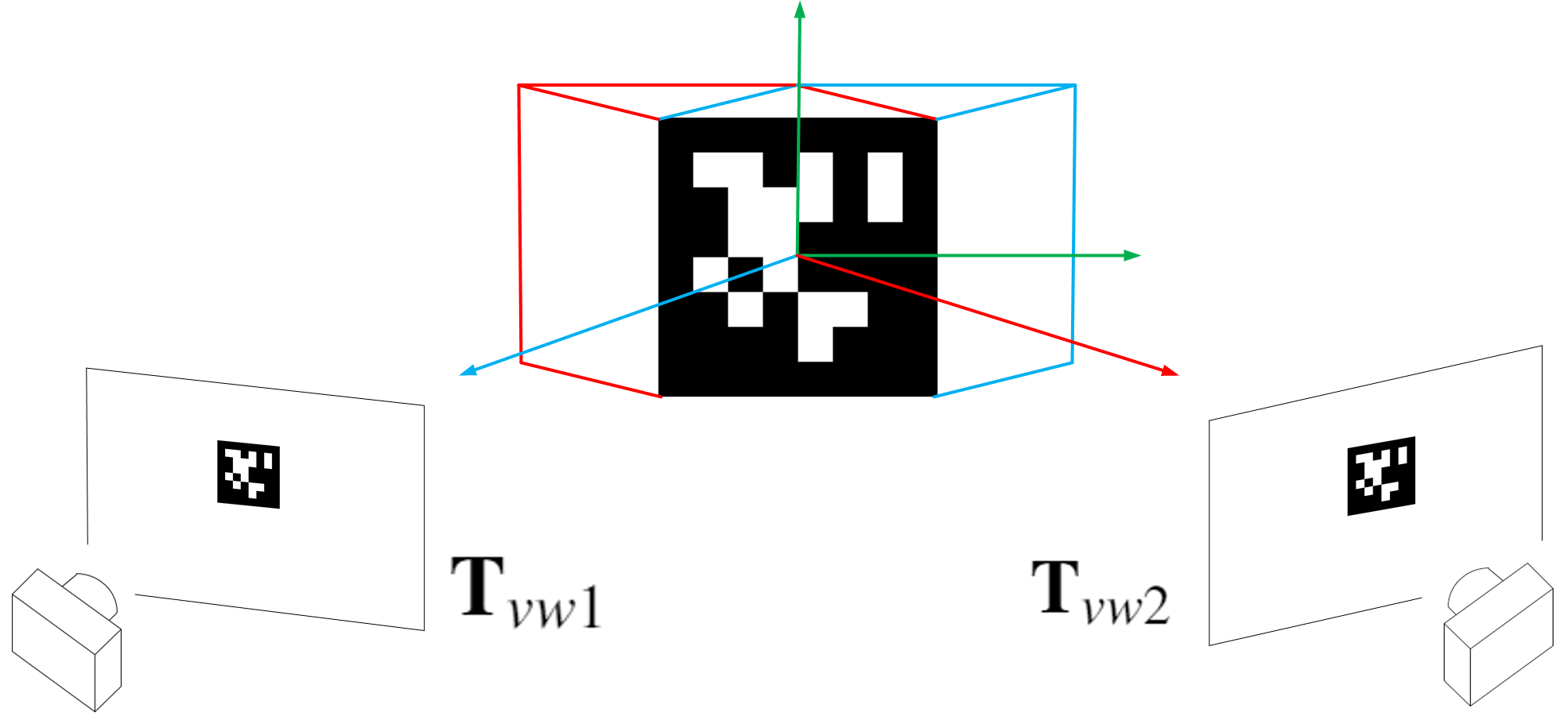}
	\caption{An illustration of the rotational ambiguity problem. When the camera is moving with the robot, the image noise shall weaken the perspective such that a marker could project at the same pixels from two different poses, $\textbf{T}_{vw1}$ and $\textbf{T}_{vw2}$. }
	\label{ambugity}
\end{figure}

Theoretically, there exists a unique estimation of the 6-DOF pose given four co-planar but non-collinear points like the corners of a marker \cite{horaud1989analytic}. Yet in the real-world, when the projected marker on the image plane is small and the relative motion between the camera and the marker is fast, the perspective effects become weak due to sensory noise. Therefore, in weak-perspective conditions, the problem of two-fold ambiguity that corresponds to an unknown reflection about the plane, and flips the z-axis in the camera's frame, arises \cite{oberkampf1996iterative}. Fig.~\ref{ambugity} shows a sketch of the rotational ambiguity problem which occurs very often in real applications \cite{munoz}.
 \par
\begin{figure}[thpb]
	\centering
	\includegraphics[width=2.5in]{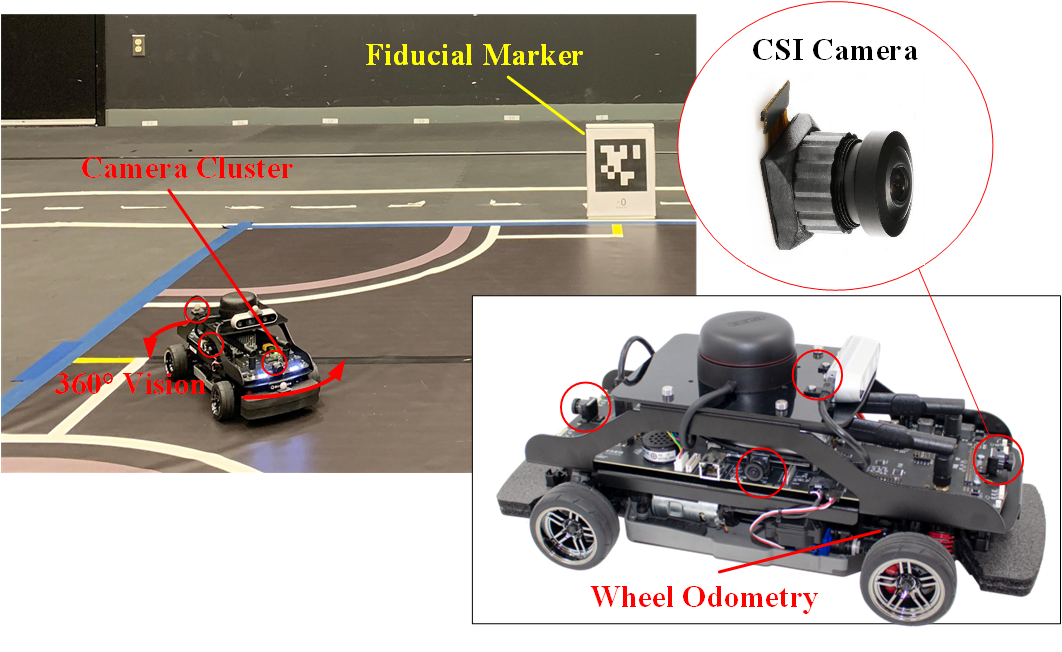}
	\caption{A diagram of our self-driving vehicle. There are four 2D CSI cameras placed on the vehicle on the front, rear, and both sides. The $360^{\circ}$ view of the camera cluster guarantees that the marker can always be observed. The on-board wheel odometer is also employed in this work.}
	\label{qcar}
\end{figure} \par
Solving the rotational ambiguity is crucial for visual servoing works where the control commands are generated based on the camera's measurements. Take our proposed framework as the example, the steering angle command mainly relies on the relative orientation between the desired waypoint and the current location of the vehicle (Fig.~\ref{qcar}). Thus, the rotational ambiguity can affect the decision of whether to turn right or left. Once the vehicle executes the wrong turning decision; the navigation task then fails. Even worse, the ambiguity of rotation leads to the flipping of translation, which ruins the positioning directly. Note that such a problem cannot be solved by methods like applying a Kalman Filter (KF) since the error is not caused by noise but by possible dual solutions. \par
In this work, we employ the AprilTag system since it has better performance over other systems \cite{wang}. Even though the rotational ambiguity is not considered in the implementation of the AprilTag system \cite{olson,wang}, the state-of-art planar pose estimation method, Infinitesimal Plane-Based Pose Estimation (IPPE) \cite{collins}, returns the two possible solutions. A straightforward method to disambiguate the two poses is comparing the reprojection errors. The pose with a lower reprojection error is then selected as the true pose. However, as shown in \cite{munoz2019} and \cite{ch2020}, the reprojection errors of the two ambiguous poses can be extremely close to each other and the one closer to the ground truth could have a higher reprojection error. Although the rotational ambiguity problem has been clarified in the novel theory \cite{collins} for marker pose estimation, it is still challenging to find the true pose from the two returned physically possible solutions. In the next section, a further survey is presented for the works that have been done for resolving rotational ambiguity in marker-based pose estimation.
\subsection{Related Work}
To the best of our knowledge, the methods for resolving the rotation ambiguity can be categorized as ambiguity detection  \cite{munoz,munoz2019}, marker improvement  \cite{tanaka2014solution,tanaka2017solving}, and additional information assistance \cite{jin,ch2020,wu,fourmy2019absolute}.\par
Munoz et al. proposed an ambiguity detection algorithm in \cite{munoz}. If the difference of the reprojection errors of the two possible poses is greater than a predetermined threshold, the pose with lower reprojection error is selected as the true pose. Otherwise, that marker detection is discarded. This algorithm is proved to be effective for marker-based SLAM \cite{munoz,munoz2019}. However, for difficult environments where only one marker is available, and also for the applications where the robot needs to respond to the camera's measurements in real-time,  simply detecting the ambiguity poses and discarding the ambiguity detections will lead to loss of localization and unstable performance. \par
Tanaka et al. \cite{tanaka2014solution,tanaka2017solving} presented a new type of marker called LentiMark by adding two new Moiré patterns to the conventional marker design. Unique angle information can be obtained from LentiMark, which eliminates the pose ambiguity. The main limitation of LentiMark is the complicated fabrication compared to that of the low-cost conventional marker.\par
Besides the methods mentioned above, additional information assistance is also very effective for mitigating the rotational ambiguity. The source of  the alleged additional information can from another sensor, predefined motion model, etc. Jin et al. \cite{jin} introduced an algorithm based on the measurements from an RGBD camera. The depth pose estimation provides optimization constraints such that the optimal pose is the one aligned with the plane in the depth space. While this algorithm requires an RGBD camera. Ch'ng et al. \cite{ch2020} developed a method that solves ambiguity by examining the consistencies of a set of markers across multiple views. This method shows good potential in marker-based SLAM, but it is not applicable for cases where only one marker is available and the robot needs to respond based on a single view. Wu et al. \cite{wu} showed a filtering based approach to disambiguate the poses. However, their approach only considers the 3D object-space error. Visual-IMU-wheel odometry \cite{lee2020} methods have the capability to deal with ambiguity poses using factor-graph based optimization. A factor-graph based approach using IMU preintegration is presented in \cite{fourmy2019absolute} to solve the rotational ambiguity. Although factor-graph is very effective as it uses the measurements from inertia sensors to construct constraint edges, the graph optimization is usually a back-end process that cannot provide the optimized results for the current frame in real-time. 
\subsection{Contributions}
We first introduce a multiple-camera model for fusing the features detected on different image planes and a KF to fuse the sensor data from the camera cluster and the wheel odometry. Then, we propose a novel algorithm to solve the rotational ambiguity by constructing a cost function, $e$, at the feature level. In particular, the cost function considers both the 2D reprojection error (image-space error), and the 3D object-space error. This algorithm is capable of providing robust pose estimation under weak-perspective conditions caused by robot motion and limited marker size. Moreover, our algorithm only requires one marker and the current frame to work. 

\section{PROPOSED FRAMEWORK}
\subsection{Multiple-Camera Model}
\begin{figure}[thpb]
	\centering
	\includegraphics[width=2.0in]{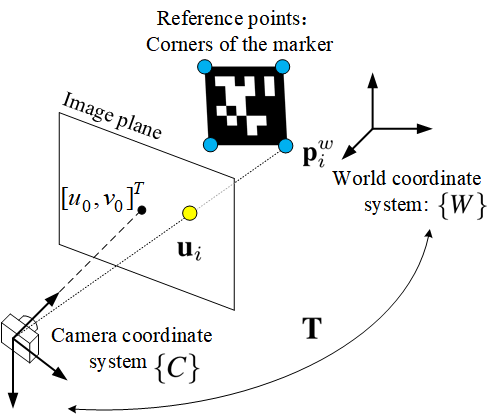}
	\caption{The pinhole camera model.  $\textbf{p}^{w}_{i}=[ x^{w}_{i},y^{w}_{i},z^{w}_{i}]^{T}$ are the 3D world coordinates of the 4 reference points. $\textbf{u}_{i}=[u_{i},v_{i}]^{T}$ are the 2D image coordinates of the projection in the image plane. $[u_0,v_0]^{T}$ is the optic center of the image plane.}
	\label{pinhole}
\end{figure}

Before introducing the coupled multiple-camera model,
we first present the pinhole camera model (see Fig.~\ref{pinhole}) for a single camera:
\begin{equation}
	s_{i} \left[\begin{array}{c} \textbf{u}_{i} \\ 1 \end{array}\right] = \textbf{K}\textbf{T}\left[\begin{array}{c} \textbf{p}^{w}_{i} \\1 \end{array}\right]\label{e1}
\end{equation}
where $s_{i}$ are scalar projective parameters for $i=\{1, 2,3, 4\}$ being the corners of the marker. $\textbf{K}\in\mathbb{R}^{3\times3}$ is the camera intrinsic matrix determined by calibration \cite{furgale}. $\textbf{T} \in\mathrm{SE(3)}$ is defined as the camera extrinsic matrix that represents the transmission from the world coordinate system $\left\{ W \right\}$ to the camera coordinate system $\left\{ C \right\}$. Detailed math preliminaries of $\mathrm{SE(3)}$ are introduced in \cite{barfoot}.
\begin{figure}[thpb]
	\centering
	\includegraphics[width=2.8in]{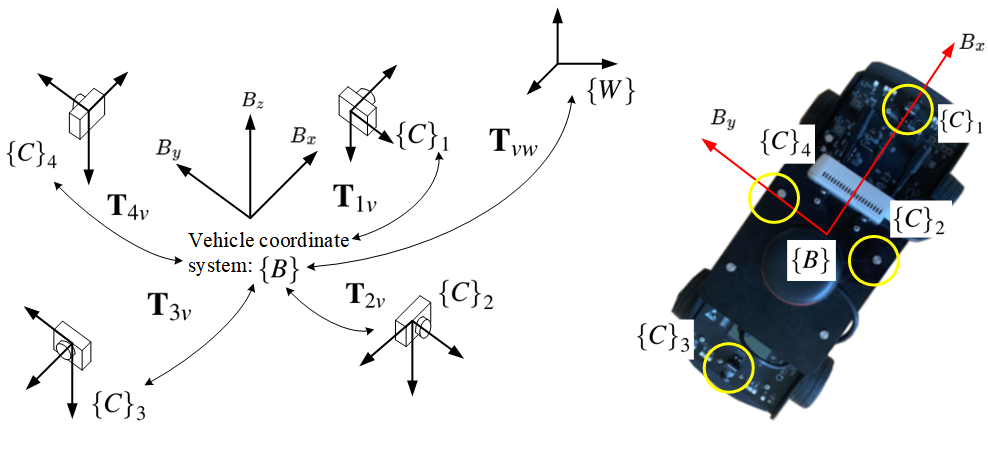}
	\caption{The multiple-camera model. $\textbf{T}_{1v}$, $\textbf{T}_{2v}$ , $\textbf{T}_{3v}$ , and $\textbf{T}_{4v}$  represent the extrinsic matrices that describe the transmission from the camera coordinate systems $\left\{ C \right\}_{1}$,  $\left\{ C \right\}_{2}$,  $\left\{ C \right\}_{3}$, and $\left\{ C \right\}_{4}$ to the vehicle coordinate system $\left\{ B \right\}$, respectively. These extrinsic matrices are determined by calibration \cite{furgale}. }
	\label{pinhole2}
\end{figure} \par
Due to the distribution of the cameras (see Fig.~\ref{pinhole2}), only the adjacent cameras have overlapping views.  Hence, no more than two cameras can observe the marker at a moment. Looking down from above, the cameras are indexed with the first at the front of the vehicle, incrementing clockwise around the vehicle. Then, we denote the index number(s) of the camera(s) that can observe the marker as  $j \subseteq \{\{ 1\},\{ 2\},\{ 3\},\{ 4\},\{ 1,2\}, \{ 2,3\},\{ 3,4\} ,\{ 4,1\}             \} $.\par
Inspired by the camera model presented in \cite{tribou}, the proposed multiple-camera model is:
\begin{equation}
	s_{ij} \left[\begin{array}{c} \textbf{u}_{ij} \\ 1 \end{array}\right] = \textbf{K}_{j}\textbf{T}_{jv}\textbf{T}_{vw}\left[\begin{array}{c} \textbf{p}^{w}_{i} \\1 \end{array}\right]\label{mcmodel}
\end{equation}
where $s_{ij}$ are scalar projective parameters and $\textbf{u}_{ij}$ is the 2D projection of the $i$th reference point on the image plane of the ${j \rm th}$ camera. $\textbf{K}_{j}$ denotes the camera intrinsic matrix of the ${j \rm th}$ camera. $\textbf{T}_{jv}$ is introduced in the caption of Fig.~\ref{pinhole2}. $\textbf{T}_{vw}$ is the extrinsic matrix that describes the transmission from the vehicle coordinate system, $\left\{ B \right\}$, to the world coordinate system, $\left\{ W \right\}$. Note that no central camera is defined here and we use the pose of the vehicle to represent the pose of the camera cluster.  In the cases where two cameras see the marker at the same time, we define the camera index numbers as $j = \{ j_{1},j_{2}\} $. The reprojection error of a given pose $\textbf{T}_{vw}$  becomes:
\begin{equation}
\begin{aligned}
& e_{1}:\mathrm{SE(3)} \rightarrow \mathbb{R}^{+}\\
& e_{1}(\textbf{T}_{vw}) = \sum_{j=j_{1}}^{j_{2}}\sum_{i=1}^4||\frac{1}{s_{ij}}\textbf{K}_{j}\textbf{T}_{jv}\textbf{T}_{vw}\left[\begin{array}{c} \textbf{p}^{w}_{i} \\1 \end{array}\right] - \left[\begin{array}{c} \textbf{u}_{m ij} \\ 1\end{array}\right]||^{2} 
\end{aligned}
\label{vresi}
\end{equation}
where  $\textbf{u}_{m ij}=[u_{m ij}, v_{m ij}]^{T}$ is the observed 2D projection of the $i\mathrm{th}$ reference point on the image plane of the $j\mathrm{th}$ camera. As for the scenes with only one camera observing the marker, the summation $\sum_{j=j_{1}}^{j_{2}}$ in Eq.~(\ref{vresi}) shall be removed.\par
The vehicle pose  $\textbf{T}_{vw}$ is determined through minimizing the reprojection error shown in Eq.~(\ref{vresi}), which is inherently a least square problem that can be solved using the Levenberg Marquardt (LM) algorithm \cite{roweis}. The initial values for LM optimization are obtained from solving IPPE \cite{collins}.
\subsection{Kinematics of the Vehicle}\label{sectionb}
\begin{figure}[thpb]
	\centering
	\includegraphics[width=1.6in]{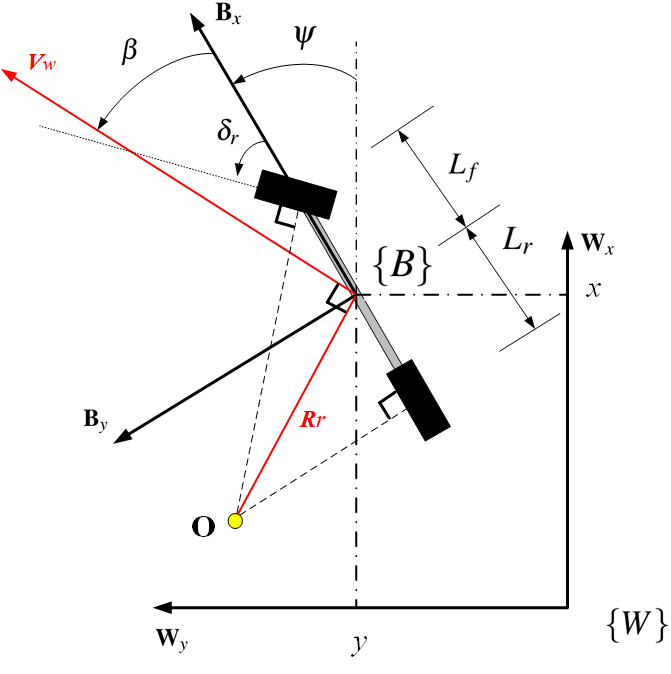}
	\caption{The 3-DOF bicycle model.  $\left\{ B \right\}$ and  $\left\{ W \right\}$ denote the vehicle and world coordinate systems, respectively. $\psi$ is the yaw orientation of the vehicle w.r.t. $\left\{ W \right\}$. $\delta_{r}$ is the steering angle of the front wheel. Depending on $\delta_{r}$, the vehicle turns about the turning point, $\textbf{O}$, with a turning radius, $R_{r}$. $\beta$ is the sideslip angle that describes the orientation of the velocity vector, $v_{w}$, w.r.t. the vehicle coordinate system. $[x,y]^{T}$ is the 2D position of the vehicle w.r.t $\left\{ W \right\}$. The centre of gravity of the vehicle is close to the centre of the vehicle chassis so that we assume that they are the same, i.e. $L_{f} = L_{r}$.}		
	\label{bicycle}
\end{figure}

The wheel odometer provides the measurement of linear velocity, $\textbf{V}_{w}$, by transforming the encoder counts. Moreover, the steering angle $\delta_{r}$ can be measured by a angle recorder in this work. To propagate the pose of the vehicle, a 3-DOF bicycle model (see Fig.~\ref{bicycle}) is employed.
 \par
From the geometric relation behind Fig.~\ref{bicycle}, the sideslip angle $\beta$ is obtained by:
\begin{equation}
	\beta = \mathrm{tan^{-1}}(\dfrac{1}{2}\mathrm{tan}\delta_{r})
	\label{beta}
\end{equation}\par 
The turning radius $R_{r}$ becomes:
\begin{equation}
	R_{r} = \dfrac{l}{\mathrm{cos}\beta~\mathrm{tan}\delta_{r}}
	\label{Rr}
\end{equation} 
where $l$ is wheelbase of the vehicle. Given the linear velocity, $v_{w}$, and turning radius, $R_{r}$, the vehicle’s angular rate is as follows:
\begin{equation}
	\dot{\psi} = \dfrac{v_{w}}{R_{r}}
\end{equation} \par
The velocity of the vehicle w.r.t. $\left\{ W \right\}$ is then:
\begin{equation}
\left[\begin{array}{c} 	v_{x}\\ v_{y}  \\\dot{\psi} \end{array}\right]=\left[\begin{array}{c} 	v_{w} \mathrm{cos}(\beta + \psi)\\ v_{w} \mathrm{sin}(\beta + \psi) \\\dot{\psi} \end{array}\right]\label{wstate}
\end{equation} \par
In this system, $\delta_{r}$ is available from the steering angle recorder and $v_{w}$ is obtained from the wheel odometer. Based on the equations introduced in this section, $[v_{x}, v_{y}, \dot{\psi} ]^{T}$ can be measured. Note that we use a different system model (see Eq.~(\ref{system})) to construct the KF in the following section for simplicity.\par
\subsection{Kalman Filter Based Sensor Fusion}\label{kalman}
A KF is employed to fuse the measurements from the camera cluster and the wheel odometer. Following \cite{kalman}, the filter state vector is defined as $	\textit{\textbf{x}}=[x,y,\psi,v_{x},v_{y},\dot{\psi}]^{T} $ and the discrete model is given as:
\begin{equation}	
	\begin{aligned}
		& \textit{\textbf{x}}_{k}=\textit{\textbf{A}}\textit{\textbf{x}}_{k-1}+\gamma_{k}\\
		& \textit{\textbf{z}}_{k}=\textit{\textbf{H}}\textit{\textbf{x}}_{k}+\upsilon_{k}\label{system}
	\end{aligned}
\end{equation}
where $\textit{\textbf{A}} = \scriptsize \left[\begin{array}{cc}\textbf{I}_{3\times3} & \textbf{dt  }\cdot\textbf{I}_{3\times3}\\\textbf{0}_{3\times3}&\textbf{I}_{3\times3} \end{array}\right] \scriptsize$ is a 6$\times$6 matrix. Here, $\textbf{dt}$ denotes the sample period, and k denotes the sample step. $\gamma_{k}$ is the systematic noise vector described by a zero-mean Gaussian distribution with covariance $\textit{\textbf{Q}}_{k}$. $\textit{\textbf{H}} = \textbf{I}_{6\times6}$ is an identity matrix. $\textit{\textbf{z}}_{k} =[z_{x},z_{y},z_{\psi},z_{vx},z_{vy},z_{\dot{\psi}}]^{T}$ is the observation vector. Note that $[z_{x},z_{y},z_{\psi}]^{T}$ is measured by the camera cluster and $[z_{vx},z_{vy},z_{\dot{\psi}}]^{T}$ is obtained from the wheel odometer and steering angle recorder (refer to section.\ref{sectionb}). $\upsilon_{k}$ is the observation noise vector described by a zero-mean Gaussian distribution with covariance $\textit{\textbf{R}}_{k}$.\par
The prediction is then given by:
\begin{equation}	
	\begin{aligned}
		& \hat{\textit{\textbf{x}}}_{k,k-1}=\textit{\textbf{A}}\hat{\textit{\textbf{x}}}_{k-1,k-1}\\
		& \textit{\textbf{P}}_{k,k-1}=\textit{\textbf{A}}\textit{\textbf{P}}_{k-1,k-1}\textit{\textbf{A}}^{T}+\textit{\textbf{Q}}_{k-1}\label{predict}
	\end{aligned}
\end{equation}
where $\textit{\textbf{P}}_{k-1,k-1}$ is a posterior covariance of the estimation error for sample step $k-1$.
$\textit{\textbf{P}}_{k,k-1}$  is the \textit{a priori} covariance of the estimation error for sample step $k$. 
The Kalman gain is obtained by:
\begin{equation}	
	\textit{\textbf{K}}_{k}=\textit{\textbf{P}}_{k,k-1}\label{gain}\textit{\textbf{H}}^{T}(\textit{\textbf{H}}\textbf{\textit{P}}_{k,k-1}\textit{\textbf{H}}^{T}+\textit{\textbf{R}}_{k})^{-1}
\end{equation}\par
The update process will be then:
\begin{equation}	
	\begin{aligned}
		& \hat{\textit{\textbf{x}}}_{k,k}=\hat{\textit{\textbf{x}}}_{k,k-1} +\textit{\textbf{K}}_{k}(\textit{\textbf{z}}_{k}-\textit{\textbf{H}}\hat{\textit{\textbf{x}}}_{k,k-1})\\
		& \textit{\textbf{P}}_{k,k}=(\textbf{I}-\textit{\textbf{K}}_{k}\textit{\textbf{H}})\textit{\textbf{P}}_{k,k-1}\label{update}
	\end{aligned}
\end{equation}
where $\textit{\textbf{P}}_{k,k}$ is the \textit{a posterior} covariance of the estimation error of sample step $k$.\par

\subsection{Resolving Rotation Ambiguity}\label{solve}
Suppose $\hat{\textit{\textbf{x}}}_{k,k-1}$ is the \textit{a priori} of sample step $k$ (see Eq.~(\ref{predict})), we defined the pose associated with $\hat{\textit{\textbf{x}}}_{k,k-1}$ as $\textbf{T}_{w}\in \mathrm{SE(3)}$. The formula that transforms the Euler angles and translation vector into $\mathrm{SE(3)}$ is shown in \cite{barfoot}. We define the function of this transmission as:
\begin{equation}
\begin{aligned}
&\textit{f}: \mathbb{R}^{6}\rightarrow\mathrm{SE(3)} \\
&\textbf{T}_{w}  = f([\hat{\textit{\textbf{x}}}_{k,k-1}^{T},\textbf{0}_{1\times 3}]^{T})
\end{aligned}
	 \label{se3}
\end{equation}
where $[\hat{\textit{\textbf{x}}}_{k,k-1}^{T},\textbf{0}_{1\times 3}]\in \mathbb{R}^{6}$  represents $\hat{\textit{\textbf{x}}}_{k,k-1}^{T}$  augmented by $\textbf{0}_{1\times 3}$. That is, we set the pitch, roll angles and $Z$ position as zero. The core of the proposed algorithm for resolving the rotational ambiguity is to minimize the value of a cost function at feature-level:
\begin{equation}
\begin{aligned}
&\textit{e}: \mathrm{SE(3)} \rightarrow \mathbb{R}^{+}\\
&e(\textbf{T}_{vw})  = e_{1}(\textbf{T}_{vw})+e_{2}(\textbf{T}_{vw})
\end{aligned}
	 \label{costfun}
\end{equation} 
where $\textbf{T}_{vw}$ is the to be determined pose of the vehicle and $e_{1}(\textbf{T}_{vw})$ is the reprojection error shown in Eq.~(\ref{vresi}). The definition of $e_{2}(\textbf{T}_{vw})$ is given by:
\begin{equation}
\begin{aligned}
&e_{2}: \mathrm{SE(3)} \rightarrow \mathbb{R}^{+}\\
&{e}_{2}(\textbf{T}_{vw})=\sum_{i=1}^4 ||\textbf{T}_{vw}\left[\begin{array}{c} \textbf{p}^{w}_{i} \\1 \end{array}\right] - \textbf{T}_{w}\left[\begin{array}{c} \textbf{p}^{w}_{i} \\1 \end{array}\right] ||^{2} \label{costfun2}
   \end{aligned}
\end{equation}
where $\textbf{T}_{w}$ is the pose associated with the \textit{a priori} from the prediction of the KF (see Eq.~(\ref{predict})). $e_{2}(\textbf{T}_{vw})$ describes the error between the 3D coordinates of two points, $\begin{scriptsize}\textbf{T}_{vw}\left[\begin{array}{c} \textbf{p}^{w}_{i} \\1 \end{array}\right]\end{scriptsize}$ and $\begin{scriptsize}\textbf{T}_{w}\left[\begin{array}{c} \textbf{p}^{w}_{i} \\1 \end{array}\right]\end{scriptsize}$, in $\left\{ B \right\}$. There is no  subtraction defined on $\mathrm{SE(3)}$ \cite{barfoot}, but we can evaluate the difference between $\textbf{T}_{w}$ and $\textbf{T}_{vw}$ through Eq.~(\ref{costfun2}). Hence, the cost $e$ is actually a sum of the 2D image-space-error $e_{1}(\textbf{T}_{vw})$ and 3D object-space error $e_{2}(\textbf{T}_{vw})$.\par
Suppose the two poses given by the IPPE method \cite{collins} are $\textbf{T}_{vw1}$ and $\textbf{T}_{vw2}$. The unambiguous pose $\textbf{T}_{vw}^{*}$ is given by:
\begin{equation}
		\begin{aligned}
		& \textbf{T}_{vw}^{*}\in \left\{ \textbf{T}_{vw1},\textbf{T}_{vw2}\right\} \label{decide}~\mathrm{s.t.}\\
		& e(\textbf{T}_{vw}^{*})=\mathrm{argmin} \left\{ e(\textbf{T}_{vw1}),e(\textbf{T}_{vw2})\right\} 
	\end{aligned}
\end{equation}\par
That is, the pose which has a smaller cost value is selected to be $\textbf{T}_{vw}^{*}$. \par
There are two issues we want to clarify for this algorithm. First, usually the direct measurements are used to construct the residual errors, such as the preintegration-based methods proposed in \cite{lee2020}. In this work, we use kinematics and integration to obtain  $\textbf{T}_{w}$, then we construct $e_{2}(\textbf{T}_{vw})$. The direct measurements are the encoder counts and steering angle. Thus, errors due to kinematics and integration are brought into $e_{2}(\textbf{T}_{vw})$. Nevertheless,  as illustrated in \cite{ch2020,wu}, the true pose is one of the two possible poses returned by the IPPE method \cite{collins}. Moreover, the true pose can be found using additional information from other sensors \cite{jin}. Hence, when solely concerning the rotation ambiguity problem, Eq.~(\ref{costfun}) is still an effective criteria. Secondly, we augmented $\hat{\textit{\textbf{x}}}_{k,k-1}\in \mathbb{R}^{3}$ into $\mathbb{R}^{6}$ to compute $\textbf{T}_{w}$, which seems superfluous for a ground vehicle. However, using $\mathrm{SE(3)}$ indicates that  Eq.~(\ref{costfun}) can also be used for aerial robots, while the kinematics and the KF will be different. 
\subsection{Lateral and Longitudinal Control}
\begin{figure}[thpb]
	\centering
	\includegraphics[width=1.6in]{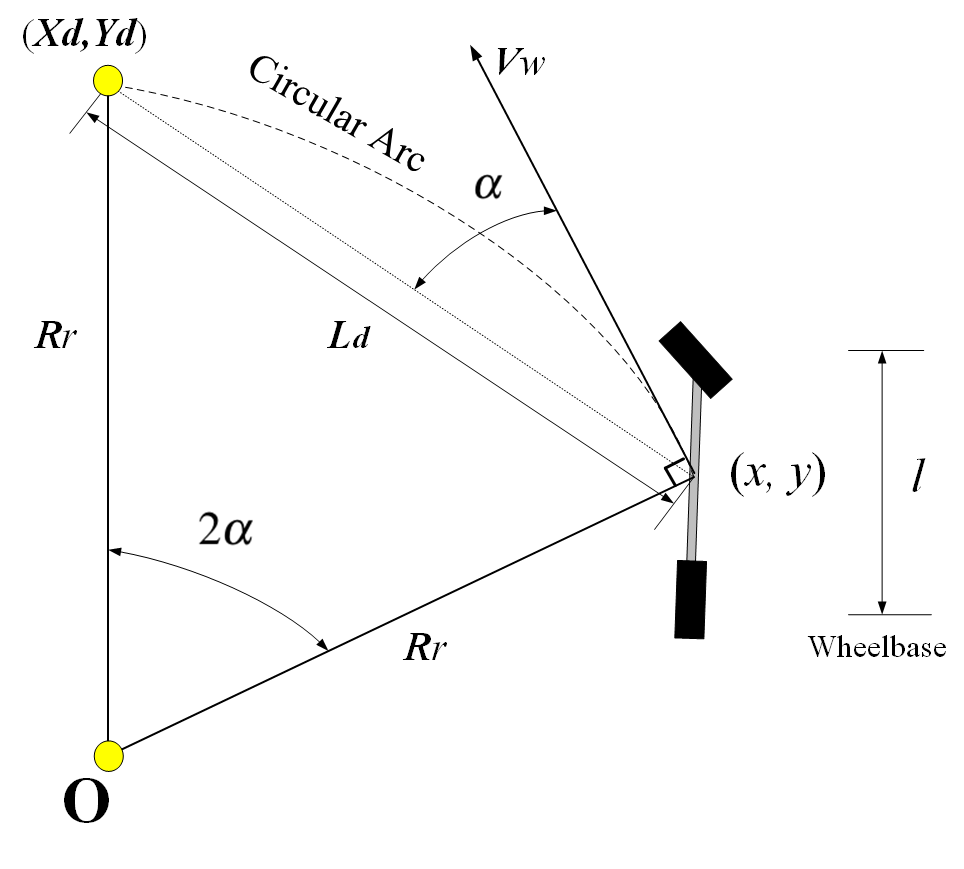}
	\caption{The Ackermann steering geometry. $(\textit{X}_{d},\textit{Y}_{d})$ is the desired waypoint. $\alpha$ is the angle between the desired waypoint and the vehicle velocity. $\textit{L}_{d} =\sqrt{\left(Y_{d}-y\right)^{2}+\left(X_{d}-x\right)^{2}}$ is the distance between the desired waypoint and the centre of the vehicle chassis. } 	
	\label{steer}
\end{figure}

The lateral controller is based on the classic Ackermann steering geometry. As illustrated in Fig.~\ref{bicycle}, the vehicle turns about the turning point $\textbf{O}$, with a turning radius $\textit{R}_{r}$. Thus, the trajectory of the centre of the vehicle chassis will be a circular arc as shown in Fig.~\ref{steer}.
 \par
At discrete control iteration $k$, from the geometry shown in Fig.~\ref{steer}, we have:
\begin{equation}	
	\textit{L}_{d,k}\mathrm{cos}\alpha_{k} = \textit{R}_{r,k}\mathrm{sin}(2\alpha_{k})
	\label{g1}
\end{equation}
where $\alpha_{k}$ is shown as follows: 
\begin{equation}	
	\alpha_{k} =\arctan (\frac{y_{k}-\textit{Y}_{d,k}}{x_{k}-\textit{X}_{d,k}})-\psi_{k}-\beta_{k}
	\label{alpha}
\end{equation}\par
Thus, given a desired waypoint $(\textit{X}_{d,k},\textit{Y}_{d,k})$, the desired steering angle $\delta_{rd,k}$ is generated by substituting Eq.~(\ref{Rr}) into Eq.~(\ref{g1}):
\begin{equation}	
\delta_{rd,k} = \arctan (\frac{2\textit{l}\mathrm{sin}\alpha_{k}}{\textit{L}_{d,k}\mathrm{cos}\beta_{k}})
\label{g2}
\end{equation}\par
We employ a PI controller for the longitudinal control. The throttle input $u_{k}$ of the vehicle is given by:
\begin{equation}	
	u_{k} = p_{1}\textit{L}_{d,k} + p_{2}\sum^{k}_{n=1} \textit{L}_{d,n},~n = \{1,\cdots,k\}
	\label{control}
\end{equation}
where $ p_{1}$ and $p_{2}$ are the proportional and integral gains, respectively. We can summarize our proposed framework in the following algorithm. \par
\begin{algorithm}[ht]
    \KwIn{a desired waypoint coordinate, $(\textit{X}_{d},\textit{Y}_{d})$}
    \KwOut{steering angle, $\delta_r$, and throttle input, $u_{k}$.}
    \vspace{3mm}
    
    Define a waypoint radius, $r$. \\
    
    \While{$\sqrt{(X_d - x)^2 + (Y_d - y)^2} > r$}{
        Get images from camera. \\
        Get AprilTag features.\cite{wang} \\
        Get both possible poses $\textit{T}_{vw1}$, and $\textit{T}_{vw2}$ using the IPPE method \cite{collins} and (\ref{vresi})\\ 
        Get the \textit{a priori} from the Kalman Prediction (\ref{predict}) \\
        Compute $T_w$ using (\ref{se3}). \\
        Compute $e(\textit{T}_{vw1})$ and $e(\textit{T}_{vw2})$ using (\ref{costfun}). \\
        Compute $\textit{T}_{vw}^*$ using (\ref{decide}). \\
        Get the \textit{a posteriori} from the Kalman update (\ref{update}). \\
        Compute $\beta$, $R_r$, $L_d$, and $\alpha$ using (\ref{beta}), (\ref{Rr}), (\ref{g1}), and (\ref{alpha}). \\
        Solve for $\delta_r$ and $\textit{u}_k$ using (\ref{g2}) and (\ref{control}). \\
        
    }
        Get next waypoint.
    \caption{Proposed Framework}
\end{algorithm}\par

\noindent\textbf{\textit{Remark 1}}: The one marker case is not an artificially difficult scenario. Even if multiple markers are set in the environment, there might be a moment where only one marker is observable. Since our framework is developed for the one-marker case, it can be easily extended to the multi-marker scenario.

\section{EXPERIMENTAL RESULTS}
\begin{figure}[thpb]
	\centering
	\includegraphics[width=3.0in]{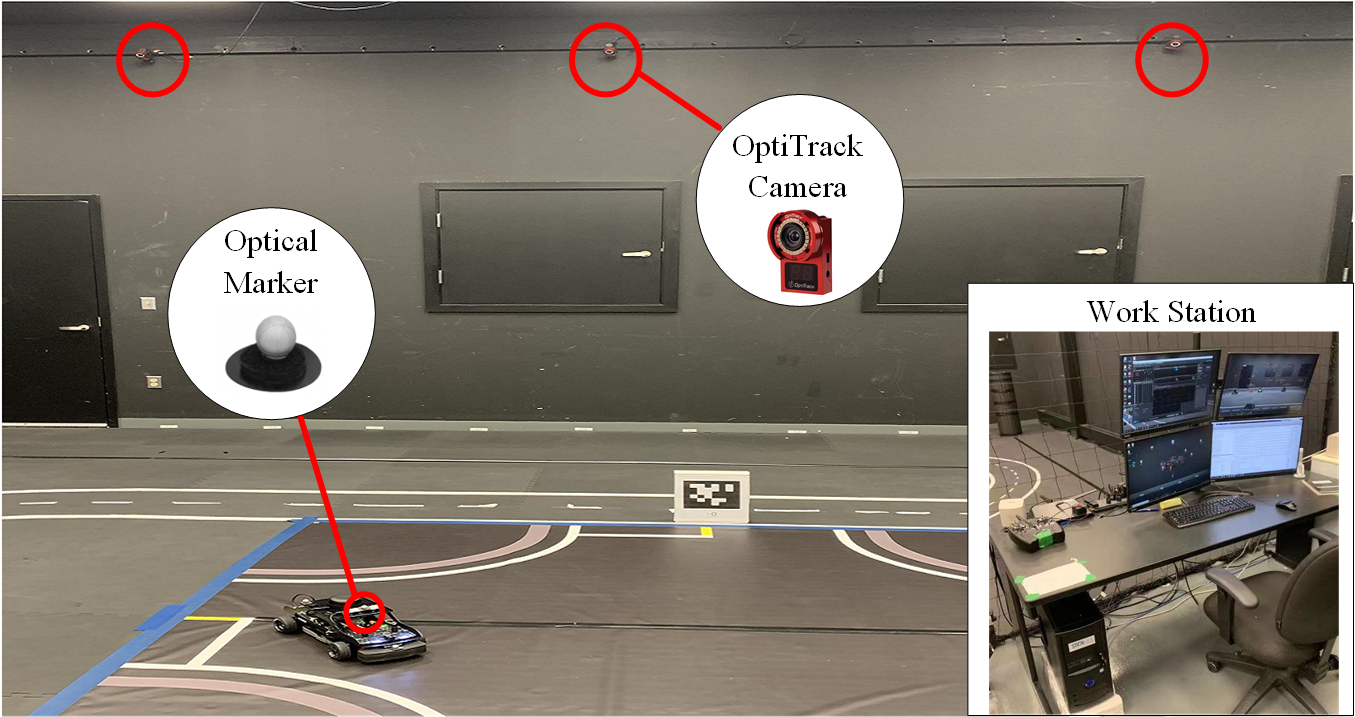}
	\caption{The OptiTrack Motion Capture System, composed of 16 OptiTrack cameras and one work station. It can track the 6-DOF pose of predefined rigid bodies via optical markers at 100 Hz.}		
	\label{opti}
\end{figure}

The proposed framework is validated using the Quanser's Qcar (see Fig.~\ref{qcar}) which is a sensor-rich autonomous vehicle for testing novel self-driving algorithms. The QCar is equipped with an NVIDIA Jetson TX2 for a computer. Our algorithm is developed using Python 3, and is implemented on an Ubuntu 18.04 OS. Quanser provides the Python libraries required to drive all the sensors and actuators. The camera cluster’s image timestamps are aligned, and the cluster can capture images while detecting the AprilTag at approximately 15 Hz. We sample from the encoder when new images are captured, and only generate the control commands after updating the KF. Overall, the proposed framework can run at approximately 11 Hz on the QCar platform. Additionally, the wheelbase of the vehicle is 25.6 cm and the steering servo can rotate from -0.5 to 0.5 rad. Our marker setup is a low-cost implementation. The marker, with a side length of 17.2 cm, is printed on A4 paper. To further verify the proposed framework, we employ the OptiTrack Motion Capture System (MoCap) at the Spacecraft Dynamics Control and Navigation Laboratory at York University to obtain the ground truth.  Fig.~\ref{opti} shows the MoCap system. \par
 \par
In the experiment, we compare the localization output of our algorithm (denoted \textbf{Our Method}), the MoCap, and the method (denoted \textbf{Method A}) which chooses the pose with lower reprojection error as the correct one. The original implementation of AprilTag \cite{wang} does not employ the IPPE method \cite{collins}, so it’s pose output is the one with the lower reprojection error. Thus,  \textbf{Method A} is the method used in the AprilTag system, which is the reason for the large measurement errors found in \cite{jin}. The vehicle is placed approximately at (200, -100) cm in $\left\{ W \right\}$, and we do not know the accurate initial poses in advance. From our experiments, we found that for this initial location, the true pose has a lower reprojection error while the vehicle is static, and the ambiguous pose has a lower reprojection error while the vehicle is moving. The objective is to navigate the vehicle through waypoints at (130, 0) cm and (50, 65) cm by turning. It should be noted that the navigation of the vehicle is based only on the output of \textbf{Our Method}, while  \textbf{Method A} runs simultaneously with \textbf{Our Method}. Moreover, we applied the same KF (Section. \ref{kalman}) for  \textbf{Method A}. The experiment video can be found at the following URL: http://3vf8.2.vu/1.\par
\begin{figure}[thpb]
	\centering
	\includegraphics[width=3.0in]{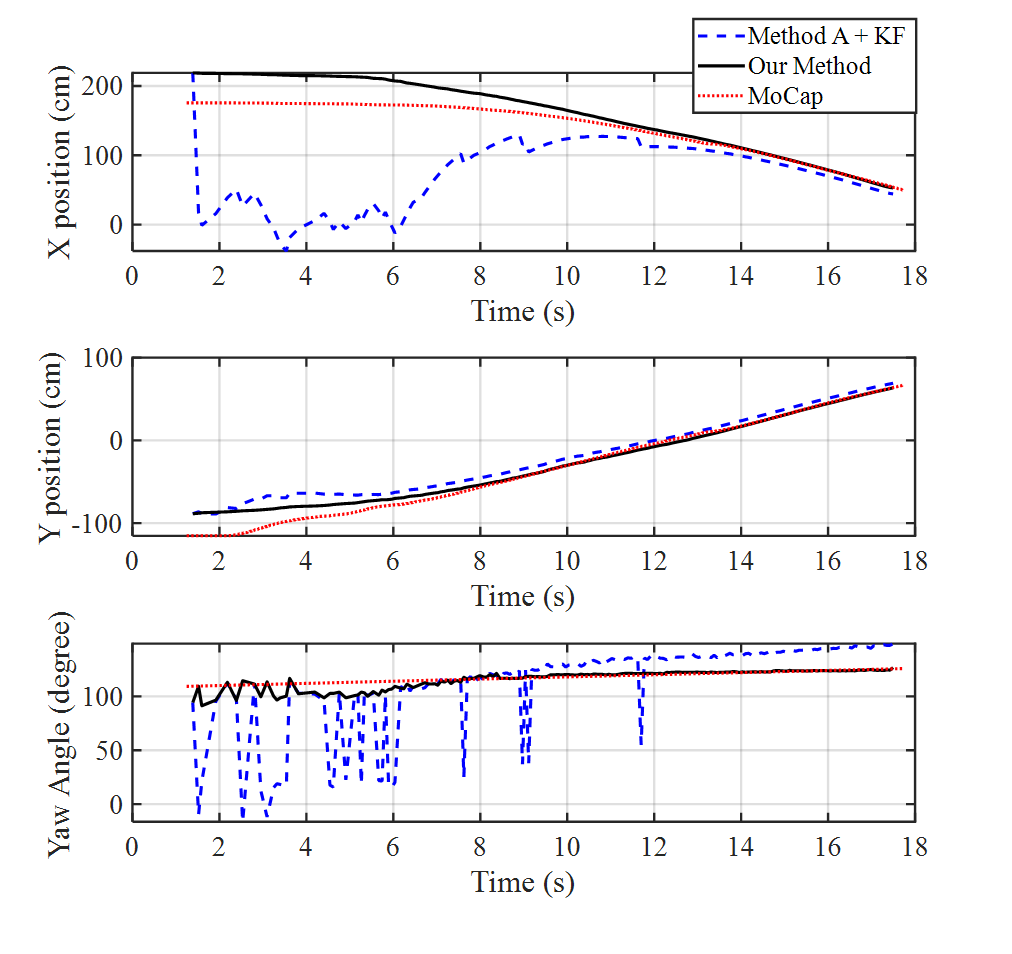}
	\caption{Comparison of the trajectories in X position, Y position, and Yaw angle along time.}		
	\label{result1}
\end{figure}
\begin{figure}[thpb]
	\centering
	\includegraphics[width=2.5in]{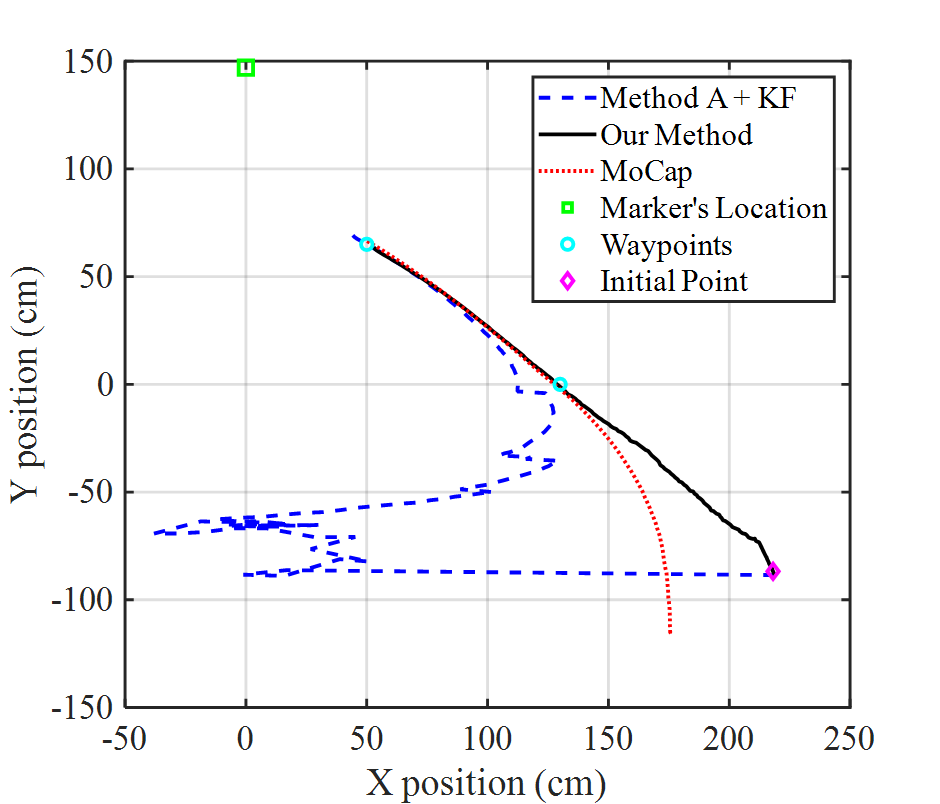}
	\caption{Comparison of trajectories in the X-Y plane of  $\left\{ W \right\}$. The four corners of the marker are set at (-8.6, 147, 31.2) cm, (8.6, 147, 31.2) cm, (8.6, 147, 14) cm, and (-8.6, 147, 14) cm in $\left\{ W \right\}$, respectively.}		
	\label{result2}
\end{figure}

As shown in Fig.~\ref{result1} and  Fig.~\ref{result2}, the outputs of  \textbf{Method A} + \textbf{KF} and  \textbf{Our Method} are almost the same at the beginning. This is because the perspective-effect is relatively strong when the vehicle is static and the true pose has the lower reprojection error. However, when the vehicle starts to move, the perspective-effect is weakened and  \textbf{Method A} + \textbf{KF}  starts to select the ambiguous pose while  \textbf{Our Method} remains stable. Therefore, without using \textbf{Our Method} to resolve ambiguity, the vehicle can execute a wrong steering command, leading to a crash. As the vehicle approaches the marker, both   \textbf{Method A}+\textbf{KF} and  \textbf{Our Method} converge to the measurement from MoCap. This is because the size of the marker in the image plane increases as the vehicle gets closer the marker thus perspective-effect is enhanced and  \textbf{Method A} + \textbf{KF}  can also work. However, an initial error exists between the output of \textbf{Our Method} and the MoCap. If the perspective-effect is insufficient due to a limited marker size, and long distance between the camera and marker, then  \textbf{Our Method} cannot handle the initial error since the cameras are the only source for the initial localization of $\left\{ W \right\}$. \par
Using the proposed framework, the vehicle successfully passes through the waypoints. Fig.~\ref{result1} and Fig.~\ref{result2} show that even if the KF is applied, such a navigation task shall fail using \textbf{Method A}. This is the default logic used in the AprilTag system for choosing the true pose from possible poses. As for the unsolved initial error problem, a deep learning-based image deblurring approach, such as HINet \cite{chen2021hinet}, is a promising solution for cases where the perspective-effect is not ideal.

\section{CONCLUSIONS}
In this work, a navigation framework of a self-driving vehicle using only one fiducial marker is proposed. The rotational ambiguity problem is solved through the computation of a novel feature level cost. The cost is computed for each pose returned by the IPPE method, using the \textit{a priori} given by the Kalman Filter, which fuses the measurements from the cameras with the wheel odometer. The pose with lower cost is selected for the Kalman Filter update and control command generation. The framework is validated experimentally and proven to be effective. Future work includes using a deep learning-based image deblurring approach to mitigate motion blur and enhance the perspective-effect, which has the potential for further improving the pose estimation accuracy.

%
%
\section*{ACKNOWLEDGMENT}

The authors would like to thank Mingfeng Yuan, Shuo Zhang, Adeel Akhtar, Shiyuan Jia, and Sahand Eivazi Adli for insightful discussions. 

\bibliographystyle{IEEEtran} 
\bibliography{reference}

\begin{thebibliography}{10}
\providecommand{\url}[1]{#1}
\csname url@samestyle\endcsname
\providecommand{\newblock}{\relax}
\providecommand{\bibinfo}[2]{#2}
\providecommand{\BIBentrySTDinterwordspacing}{\spaceskip=0pt\relax}
\providecommand{\BIBentryALTinterwordstretchfactor}{4}
\providecommand{\BIBentryALTinterwordspacing}{\spaceskip=\fontdimen2\font plus
\BIBentryALTinterwordstretchfactor\fontdimen3\font minus
  \fontdimen4\font\relax}
\providecommand{\BIBforeignlanguage}[2]{{%
\expandafter\ifx\csname l@#1\endcsname\relax
\typeout{** WARNING: IEEEtran.bst: No hyphenation pattern has been}%
\typeout{** loaded for the language `#1'. Using the pattern for}%
\typeout{** the default language instead.}%
\else
\language=\csname l@#1\endcsname
\fi
#2}}
\providecommand{\BIBdecl}{\relax}
\BIBdecl

\bibitem{olson}
E.~Olson, ``Apriltag: A robust and flexible visual fiducial system,'' in
  \emph{2011 IEEE International Conference on Robotics and Automation}.\hskip
  1em plus 0.5em minus 0.4em\relax IEEE, 2011, pp. 3400--3407.

\bibitem{wang}
J.~Wang and E.~Olson, ``Apriltag 2: Efficient and robust fiducial detection,''
  in \emph{2016 IEEE/RSJ International Conference on Intelligent Robots and
  Systems (IROS)}.\hskip 1em plus 0.5em minus 0.4em\relax IEEE, 2016, pp.
  4193--4198.

\bibitem{ch2020}
S.-F. Ch’ng, N.~Sogi, P.~Purkait, T.-J. Chin, and K.~Fukui, ``Resolving
  marker pose ambiguity by robust rotation averaging with clique constraints,''
  in \emph{2020 IEEE International Conference on Robotics and Automation
  (ICRA)}.\hskip 1em plus 0.5em minus 0.4em\relax IEEE, 2020, pp. 9680--9686.

\bibitem{borowczyk}
A.~Borowczyk, D.-T. Nguyen, A.~P.-V. Nguyen, D.~Q. Nguyen, D.~Saussi{\'e}, and
  J.~Le~Ny, ``Autonomous landing of a quadcopter on a high-speed ground
  vehicle,'' \emph{Journal of Guidance, Control, and Dynamics}, vol.~40, no.~9,
  pp. 2378--2385, 2017.

\bibitem{munoz}
R.~Mu{\~n}oz-Salinas, M.~J. Mar{\'\i}n-Jimenez, E.~Yeguas-Bolivar, and
  R.~Medina-Carnicer, ``Mapping and localization from planar markers,''
  \emph{Pattern Recognition}, vol.~73, pp. 158--171, 2018.

\bibitem{munoz2019}
R.~Munoz-Salinas, M.~J. Marin-Jimenez, and R.~Medina-Carnicer, ``Spm-slam:
  Simultaneous localization and mapping with squared planar markers,''
  \emph{Pattern Recognition}, vol.~86, pp. 156--171, 2019.

\bibitem{horaud1989analytic}
R.~Horaud, B.~Conio, O.~Leboulleux, and B.~Lacolle, ``An analytic solution for
  the perspective 4-point problem,'' \emph{Computer Vision, Graphics, and Image
  Processing}, vol.~47, no.~1, pp. 33--44, 1989.

\bibitem{oberkampf1996iterative}
D.~Oberkampf, D.~F. DeMenthon, and L.~S. Davis, ``Iterative pose estimation
  using coplanar feature points,'' \emph{Computer Vision and Image
  Understanding}, vol.~63, no.~3, pp. 495--511, 1996.

\bibitem{collins}
T.~Collins and A.~Bartoli, ``Infinitesimal plane-based pose estimation,''
  \emph{International journal of computer vision}, vol. 109, no.~3, pp.
  252--286, 2014.

\bibitem{tanaka2014solution}
H.~Tanaka, Y.~Sumi, and Y.~Matsumoto, ``A solution to pose ambiguity of visual
  markers using moire patterns,'' in \emph{2014 IEEE/RSJ International
  Conference on Intelligent Robots and Systems}.\hskip 1em plus 0.5em minus
  0.4em\relax IEEE, 2014, pp. 3129--3134.

\bibitem{tanaka2017solving}
H.~Tanaka, K.~Ogata, and Y.~Matsumoto, ``Solving pose ambiguity of planar
  visual marker by wavelike two-tone patterns,'' in \emph{2017 IEEE/RSJ
  International Conference on Intelligent Robots and Systems (IROS)}.\hskip 1em
  plus 0.5em minus 0.4em\relax IEEE, 2017, pp. 568--573.

\bibitem{jin}
P.~Jin, P.~Matikainen, and S.~S. Srinivasa, ``Sensor fusion for fiducial tags:
  Highly robust pose estimation from single frame rgbd,'' in \emph{2017
  IEEE/RSJ International Conference on Intelligent Robots and Systems
  (IROS)}.\hskip 1em plus 0.5em minus 0.4em\relax IEEE, 2017, pp. 5770--5776.

\bibitem{wu}
P.-C. Wu, J.-H. Lai, J.-L. Wu, and S.-Y. Chien, ``Stable pose estimation with a
  motion model in real-time application,'' in \emph{2012 IEEE International
  Conference on Multimedia and Expo}.\hskip 1em plus 0.5em minus 0.4em\relax
  IEEE, 2012, pp. 314--319.

\bibitem{fourmy2019absolute}
M.~Fourmy, D.~Atchuthan, N.~Mansard, J.~Sola, and T.~Flayols, ``Absolute
  humanoid localization and mapping based on imu lie group and fiducial
  markers,'' in \emph{2019 IEEE-RAS 19th International Conference on Humanoid
  Robots (Humanoids)}.\hskip 1em plus 0.5em minus 0.4em\relax IEEE, 2019, pp.
  237--243.

\bibitem{lee2020}
W.~Lee, K.~Eckenhoff, Y.~Yang, P.~Geneva, and G.~Huang, ``Visual-inertial-wheel
  odometry with online calibration,'' in \emph{Proceedings of the 2020
  International Conference on Intelligent Robots and Systems (IROS), Las Vegas,
  NV, USA}, 2020, pp. 24--30.

\bibitem{furgale}
P.~Furgale, J.~Rehder, and R.~Siegwart, ``Unified temporal and spatial
  calibration for multi-sensor systems,'' in \emph{2013 IEEE/RSJ International
  Conference on Intelligent Robots and Systems}.\hskip 1em plus 0.5em minus
  0.4em\relax IEEE, 2013, pp. 1280--1286.

\bibitem{barfoot}
T.~D. Barfoot, \emph{State estimation for robotics}.\hskip 1em plus 0.5em minus
  0.4em\relax Cambridge University Press, 2017.

\bibitem{tribou}
M.~J. Tribou, A.~Harmat, D.~W. Wang, I.~Sharf, and S.~L. Waslander,
  ``Multi-camera parallel tracking and mapping with non-overlapping fields of
  view,'' \emph{The International Journal of Robotics Research}, vol.~34,
  no.~12, pp. 1480--1500, 2015.

\bibitem{roweis}
S.~Roweis, ``Levenberg-marquardt optimization,'' \emph{Notes, University Of
  Toronto}, 1996.

\bibitem{kalman}
F.~Janabi-Sharifi and M.~Marey, ``A kalman-filter-based method for pose
  estimation in visual servoing,'' \emph{IEEE transactions on Robotics},
  vol.~26, no.~5, pp. 939--947, 2010.

\bibitem{chen2021hinet}
L.~Chen, X.~Lu, J.~Zhang, X.~Chu, and C.~Chen, ``Hinet: Half instance
  normalization network for image restoration,'' in \emph{IEEE/CVF Conference
  on Computer Vision and Pattern Recognition Workshops}, 2021.

\end{thebibliography}

\end{document}